\documentclass[conference,a4paper]{IEEEtran}

\usepackage[T1]{fontenc}
\usepackage{microtype}
\usepackage{booktabs}
\usepackage{array}
\usepackage{siunitx}
\usepackage{amsmath}
\usepackage[table]{xcolor}
\usepackage{graphicx}
\usepackage[hidelinks]{hyperref}
\usepackage{balance}
\input{orcid-logo}

\newcommand{\PS}{Poll--Schumann}

\title{Extending Aircraft Coverage of the Poll--Schumann Model Using Public Data}

\author{
\IEEEauthorblockN{Gabriel Jarry\,\orcid{0000-0002-7294-4217},
Ramon Dalmau\,\orcid{0000-0003-3587-7331},
Philippe Very}
\IEEEauthorblockA{Aviation Sustainability Unit, EUROCONTROL Innovation Hub\\
Brétigny-sur-Orge, France\\
\href{mailto:gabriel.jarry@eurocontrol.int}{gabriel.jarry@eurocontrol.int}}
}

\begin{document}
\raggedbottom
\maketitle

\begin{abstract}
The Poll--Schumann model requires airframe--engine parameters, but published tables cover only part of the fleet. Fuel-burn estimates for an unrepresented type require type-specific coefficients or a substitute. We reconstruct parameters using geometry from certification documents and manufacturer manuals, together with ICAO engine records. Published equations supply the physical parameters. When the propulsion-efficiency scale $\eta_1$ is unavailable for a configuration, we estimate it from public aircraft and engine characteristics using a regression fitted to published scales. The inventory covers 90 type designators, including 22 absent from the direct pycontrails table and 18 without a listed substitute. Across 68 types, mean absolute percentage error (MAPE) against pycontrails was 7.38\% for drag and 3.14\% for design Mach. Across 66 types in 21 families, efficiency regression achieved 3.91\% MAPE with the evaluated family excluded from input selection and fitting. On 100 A320-214 flights, both CFM56-5B4/P parameter sets used common airframe geometry and fuel-flow limits. Cruise fuel flow MAPE rose from 4.63\% with the published set to 5.03\% with the reconstructed set. Integrated-fuel error over covered airborne intervals rose from 6.86\% to 8.18\%. Public technical records extend Poll--Schumann parameter coverage. On the evaluated flights the reconstructed set gives higher errors for cruise flow and covered-airborne fuel.
\end{abstract}


\section{Introduction}

Aircraft performance models support trajectory prediction and fuel and emissions estimates for air traffic management (ATM). Applying these models requires parameters that describe the aircraft and its engines. The \PS{} model provides physical equations and published parameter tables~\cite{poll2020part1,poll2020part2}, but those tables cover only part of the fleet. For types without a direct entry, trajectory-modelling tools need reconstructed coefficients or a documented substitute. We use pycontrails as a source of reference coefficients and type-to-type substitutions~\cite{marc_shapiro_2026_19932206}.

Public inputs come from type-certificate data sheets for aircraft variants and certified limits~\cite{easaA064}, technical publications for geometry~\cite{icaoC919Airport}, and a databank of certified engine characteristics~\cite{icaoedb32}. However, the required values are rarely available together for the exact airframe and engine variant. Some inputs must be approximated. Overall propulsion efficiency, the fraction of fuel power converted into propulsive power, requires an additional fitted parameter. The practical question is therefore how far these sources can extend model coverage, and what accuracy the resulting parameters provide.

We calculate physical coefficients and fit the efficiency scale from aircraft and engine records to reconstruct \PS{} parameters. We evaluate agreement with published coefficients on known types and examine the reliability of the fitted efficiency component. We then compare the complete published and reconstructed parameter sets against measured fuel consumption on recorded A320-214 flights. The contributions are an expanded parameter inventory and a method to estimate the propulsion-efficiency scale $\eta_1$ for configurations without a published value. The regression learns from published values and uses technical characteristics available for the target configuration. We measure coverage against the pinned pycontrails tables. We reconstruct already tabulated types to evaluate the method and treat their reference coefficients as benchmarks.

\section{Related Work}
\label{sec:soa}

Aircraft performance models provide different ways to estimate thrust, drag and fuel consumption. The Base of Aircraft Data (BADA) supports ATM applications through licensed aircraft models~\cite{nuic2010bada}. OpenAP combines public aircraft information and surveillance data in an open-source model~\cite{sun2020openap}. Poll and Schumann describe aircraft performance through dimensionless relations and characteristic parameters~\cite{poll2020part1,poll2020part2}. Subsequent developments address optimum flight conditions, engine performance and complete flight profiles~\cite{poll2024optimum,poll2024engine,poll2025part3}. The reference tables are distributed with pycontrails~\cite{marc_shapiro_2026_19932206}. We execute the model with OpenLeaf~\cite{openleaf2026}.

Missing model information can be estimated from technical records or observed flights. Part~2 of Poll and Schumann's fuel-burn estimation method already derives characteristic aircraft parameters using open-literature correlations~\cite{poll2020part2}. Sun et al. estimate drag polars from open surveillance observations using a stochastic total-energy model~\cite{sun2020dragpolar}. Alligier et al. infer mass and thrust settings from past trajectory observations to improve climb prediction~\cite{alligier2013mass}. These approaches address different unknowns: aircraft characteristics in the first two cases, and quantities that vary from flight to flight in the latter.

Other methods estimate fuel consumption from observed flights. Chatterji combines flight-track reconstruction with aerodynamic calculations and the BADA fuel flow model, validating the estimates against Flight Data Recorder measurements from an FAA test flight~\cite{chatterji2011fuel}. Seymour et al. estimate commercial aviation fuel consumption using an open-data approach~\cite{seymour2020fuel}. Chati and Balakrishnan use Gaussian-process regression to predict fuel flow and its uncertainty from operational recordings~\cite{chati2017gpr}. Jarry et al. train supervised fuel flow regression models on Quick Access Recorder (QAR) data for application to Automatic Dependent Surveillance--Broadcast (ADS-B) trajectories~\cite{jarry2024towards}. Their models combine flight variables with aircraft characteristics to support prediction across types. Subsequent work examines whether learned fuel flow models transfer to aircraft absent from training~\cite{jarry2024generalization}. Physics-guided learning provides another approach, using physical relationships to constrain fuel predictions~\cite{uzun2021physics}. These methods estimate the output of interest, whereas parameter reconstruction supplies quantities that can be reused within a physical performance model. This allows expansion where flight observations are scarce, at the cost of dependence on technical-source quality and the model assumptions.

Our work extends the application of the published \PS{} equations by assembling parameter sets from independently sourced aircraft and engine records. We assess both parameter agreement and the fuel predictions obtained from those sets. The fuel comparison runs both parameter sets in our OpenLeaf implementation~\cite{openleaf2026} on identical recorded flight states.

\section{Reconstructing Parameters from Public Data}
\label{sec:method}

We reconstructed aircraft-specific \PS{} parameters from public information on airframe geometry and engine characteristics. Most parameters were calculated using the published model equations, while the efficiency scale $\eta_1$ was estimated by regression against published values. Overall efficiency determines fuel flow for a given thrust and speed; it also affects the temperature threshold for contrail formation~\cite{schumann2000efficiency}. Table~\ref{tab:notation} summarises the parameters and their roles. Figure~\ref{fig:boundary} distinguishes the physical calculations from the fitted component.

\begin{table}[t]
\caption{Main parameters (all dimensionless, unit 1) and their physical roles~\cite{poll2024optimum,poll2024engine}. Parameters are calculated from public inputs using the model equations, except $\eta_1$, which is fitted, and $\psi_1$, which also depends on that fit. DO denotes the design optimum.}
\label{tab:notation}
\centering\small
\begin{tabular}{@{}lp{0.70\columnwidth}@{}}
\toprule
Symbol & Physical role \\
\midrule
$\psi_0$ & Zero-lift drag scale \\
$\psi_1$ & Range scale, including propulsion efficiency \\
$\psi_2,\ \psi_3$ & Lift and lift-to-drag scales, respectively \\
$\psi_4=M_{\mathrm{DO}}$ & Design-optimum Mach number \\
$\psi_5,\ \psi_6$ & Reynolds-number and lift scales, respectively \\
$\psi_7$ & Optimum-mass scale at the tropopause \\
$\tau$ & Sensitivity of span efficiency to skin friction \\
$\eta_1,\ \eta_2$ & Propulsion-efficiency scale and Mach exponent \\
$J_1,\ J_2$ & Wave-drag parameters \\
$C_{L,\mathrm{DO}}$ & Lift coefficient at the design optimum \\
$M_{\mathrm{EC}}$ & Engine characteristic Mach number \\
$T_{\mathrm{REC}}$ & Turbine-entry to freestream total-temperature ratio at maximum overall efficiency \\
\bottomrule
\end{tabular}
\end{table}

\subsection{Public aircraft and engine information}

\begin{figure}[t]
\centering
\includegraphics[width=\linewidth]{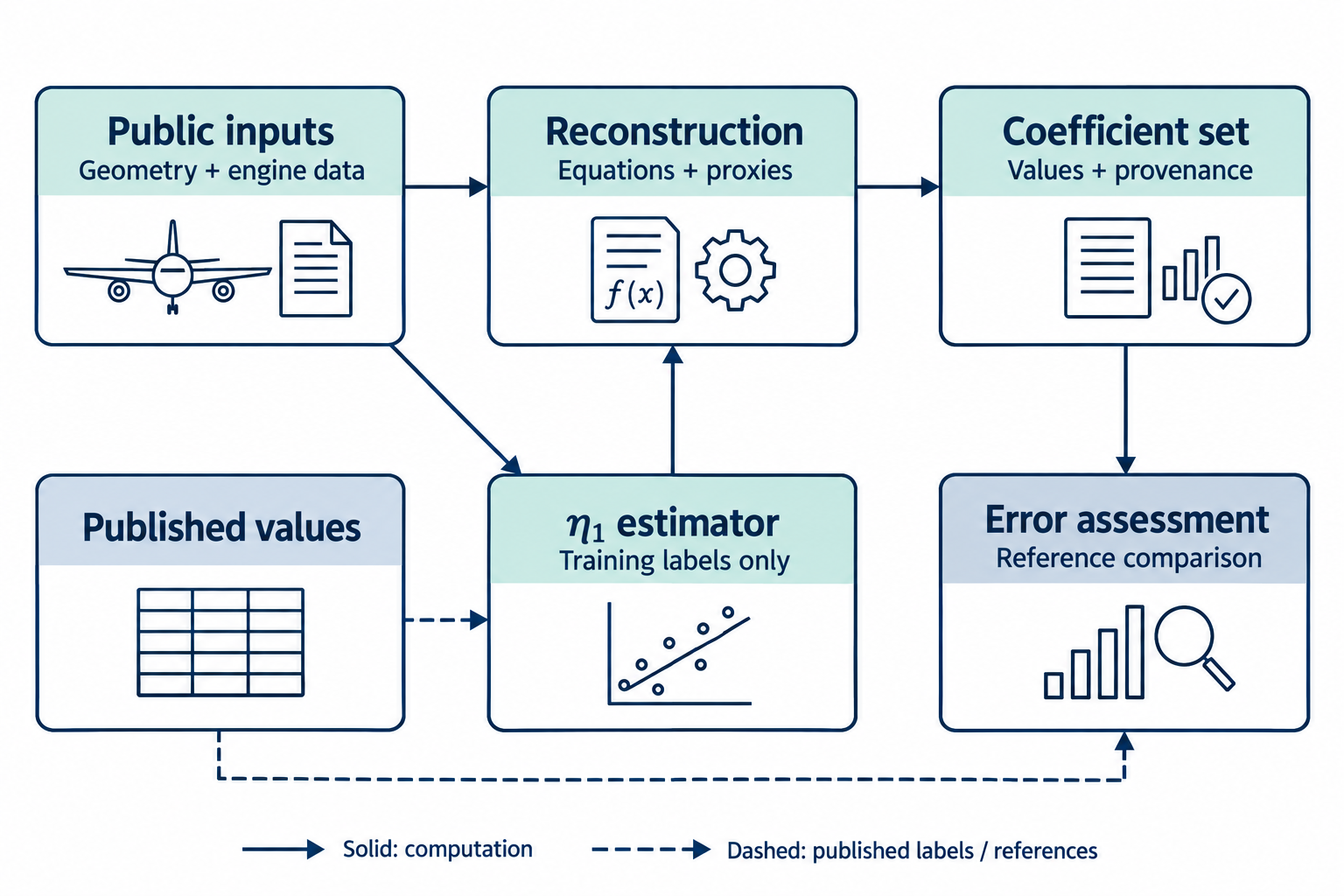}
\caption{Inputs to the reconstruction. Solid arrows show the calculations and their inputs or outputs. Dashed arrows carry published values used for fitting or comparison. The regression estimates the efficiency parameter $\eta_1$ from public predictors, using published efficiency values as training labels.}
\label{fig:boundary}
\end{figure}

Figure~\ref{fig:boundary} begins with source records for the airframe and engine.

Airframe data came from type-certificate data sheets and manufacturer airport-planning manuals. They described dimensions, wing geometry, certified masses and operating limits. Engine characteristics came from version~32 of the ICAO Aircraft Engine Emissions Databank~\cite{icaoedb32}, including pressure ratio, bypass ratio, rated thrust and take-off fuel flow. We retained the source of each value. Published technical reports supplement geometry where needed~\cite{nasaA320Geometry}. Missing quantities were estimated from dimensioned aircraft drawings in manufacturer manuals, related variants or analytical approximations, with the approximation recorded alongside the value. The C919 wingspan of 35.8\,m is reported in the ICAO working paper~\cite[Table~1]{icaoC919Airport}. For the LEAP-1C28, the ICAO EDB entry 08P28CM150 gives a bypass ratio of 10.56 (rounded)~\cite{icaoedb32}. Wetted areas derived from drawings are recorded as estimates.

We identified airframe variants and their documented engine installations from public technical sources. The suffix -214 identifies a certified A320 airframe variant, distinct from the -232 variant~\cite{easaA064}. The variant and the applicable engine databank UID jointly identify an inventory record. Each engine model was linked to the applicable records in the ICAO Engine Emissions Databank. Multiple databank records may describe the same engine model; their number therefore does not represent the number of distinct aircraft variants. An ICAO type groups airframe variants. An installation pairs a variant with a documented engine. An inventory record links that variant to an applicable databank entry. We record document editions and applicable modifications. A certified dimension may depend on whether a particular modification applies~\cite{easaA064}.

\subsection{Calculating the physical parameters}
\label{sec:physical-method}

The physical-calculation branch of Figure~\ref{fig:boundary} uses geometry and operating limits. The summary below lists its main approximations.

We used the parameter equations in Part~2 of Poll and Schumann's fuel-burn estimation method~\cite{poll2020part2}, with their subsequent aerodynamic and engine developments~\cite{poll2024optimum,poll2024engine}. The connection to fuel consumption follows the overall-efficiency definition~\cite{poll2024optimum}:
\begin{equation}
\dot m_f=\frac{F_n V}{\eta_o\,L_{\mathrm{CV}}},
\label{eq:fuel-physical}
\end{equation}
where $\dot m_f$ is total fuel flow, $F_n$ total net thrust, $V$ true airspeed, $\eta_o$ overall efficiency and $L_{\mathrm{CV}}$ fuel lower calorific value. Reconstructing drag and efficiency therefore supplies the quantities needed to predict fuel flow at a given flight state.

The calculation starts from aircraft geometry and an estimate of cruise Mach. It determines drag and the design-optimum flight condition used to define the characteristic parameters. These parameters describe the aircraft; the fuel calculation combines them with the measured mass, speed and atmospheric conditions of each flight.

The first step estimated the wetted area, meaning the external surface exposed to the airflow, for the fuselage, wings, tails and nacelles. These areas supplied the zero-lift drag parameter $\psi_0$, with shape and interference corrections from transport-aircraft design methods~\cite{obert2009,raymer2018,torenbeek1982}. We retained the empirical corrections in the \PS{} formulation, including its technology correction where applicable. Thus, independently sourced geometry enters equations that already contain empirical calibration.

The next step estimated the design-optimum Mach number, $\psi_4=M_{\mathrm{DO}}$. When an independent source explicitly reported a long-range cruise Mach number, $M_{\mathrm{LRC}}$, we used
\begin{equation}
M_{\mathrm{DO}}\approx\frac{M_{\mathrm{LRC}}}{1.035}.
\label{eq:mach-proxy}
\end{equation}
This proxy follows the approximate 3.5\% separation between long-range cruise and minimum-fuel Mach discussed by Poll and Schumann~\cite[Sec.~6]{poll2024optimum}. It also approximates the operating optimum by the design optimum; a generic quoted cruise speed was insufficient for this purpose. When no suitable long-range cruise value was available, we used an analytical estimate combining the optimum-condition equations with a crest-Mach approximation~\cite{poll2024optimum,poll2025part3}.

At that Mach number, we followed the design-optimum iteration of Poll and Schumann~\cite[pp.~1088--1089]{poll2024optimum}. An initial drag estimate determined whether to use the tropospheric or stratospheric equations; this choice remained fixed during the iteration. Each updated lift coefficient determined pressure and Reynolds number, then skin friction and drag. For aircraft with winglets, the span-efficiency correction was also included consistently in the drag derivatives. Errors in geometry or design Mach can therefore influence several output parameters. The design mass is $m_{\mathrm{des}}=0.80\,\mathrm{MTOM}_{\mathrm{ref}}$. The wing reference area $S_{\mathrm{ref}}$ and full $\mathrm{MTOM}_{\mathrm{ref}}$ enter $\psi_6$ (Eq.~58 in~\cite{poll2024optimum}). Lift balance gives $p_{\mathrm{DO}}/p_{\mathrm{TP}}=0.80\psi_6/C_{L,\mathrm{DO}}$. The ISA pressure--altitude relation then gives the design flight level for that installation.

The coefficients in Table~\ref{tab:notation} summarise this physical description. For Table~\ref{tab:coverage}, $\psi_0,\psi_2,\psi_3,\psi_4,\psi_5,\psi_6,\psi_7$ correspond to Eqs.~(16), (60), (66), (54), (56), (58), and (62), respectively, and $\tau$ to Eq.~(C-4) in~\cite{poll2024optimum}. For the engine terms, Eq.~\eqref{eq:ridge-fit} defines the fit used to predict $\eta_1$ and the bypass-ratio correlation gives $\eta_2$~\cite[Eq.~(24)]{poll2024engine}.

\par\smallskip\noindent
{\small
\begin{tabular}{@{}p{0.58\columnwidth}p{0.35\columnwidth}@{}}
\toprule
Approximation & Main quantities affected \\
\midrule
Wetted areas from drawings and shape factors & Zero-lift drag, design optimum \\
Missing dimensions from documented related variants or estimates & Wing loading, drag and efficiency inputs \\
Long-range-cruise proxy or analytical crest-Mach closure & Design Mach and dependent coefficients \\
Ridge fit to published efficiency scales & $\eta_1$ and $\psi_1$ \\
\bottomrule
\end{tabular}
}\par

\subsection{Estimating the efficiency parameter}
\label{sec:efficiency-method}

Overall efficiency also depends on the constant $\eta_1$, fitted to the published type-level efficiency values in the Poll--Schumann Part~3 engine table~\cite{poll2025part3}. This is the regression branch of Figure~\ref{fig:boundary}. Its inputs were overall pressure ratio, rated sea-level static thrust, take-off fuel flow and wing reference area. The regression used four terms: take-off fuel flow, the logarithm of pressure ratio, fuel flow divided by rated thrust, and the logarithm of wing area. Thrust, fuel flow and area were expressed in kN, kg/s and m$^2$, respectively.

For published targets $y_i$, weighted-standardised predictors $z_i$, and weights $w_i$, we minimise 
\begin{equation}
  \sum_i w_i(y_i-b-z_i^\mathsf{T}\beta)^2+\alpha\lVert\beta\rVert_2^2 .
\label{eq:ridge-fit}
\end{equation} We set $\alpha=1$ and leave $b$ unpenalised. For a target with standardised predictors $z$, the fitted model predicts $\widehat{\eta}_1=\widehat{b}+z^\mathsf{T}\widehat{\beta}$. Standardisation uses the weighted training mean and variance in each fit. Weighting gave each aircraft family equal importance, then each type within a family and each distinct input combination within a type. Duplicate installation records therefore did not receive extra weight. The weights summed to the number of training observations. The fitted $\eta_1$ then supplied design-point overall efficiency, $\eta_{o,\mathrm{DO}}$, and the range parameter $\psi_1$ (DO denotes the design optimum) through the \PS{} equations.

For the inventory and recorded-flight comparison, we use one frozen regression fitted to 167 distinct type--predictor combinations across 26 ICAO types in nine families. This fit supplies $\eta_1$ for every reconstructed record, including types with published values. We evaluated its ability to estimate efficiency across aircraft families and examined how selecting its inputs affects the estimated error (Section~\ref{sec:efficiency-evaluation}).

\subsection{The resulting parameter inventory}
\label{sec:coverage}

The \textit{Coefficient set} in Figure~\ref{fig:boundary} contains 928 complete records for 90 ICAO type designators. The design-Mach iteration failed for two A340-213 records, CFM56-5C4 (UID 2CM015) and CFM56-5C4/P (UID 7CM047). Other A342 variants remained available. Table~\ref{tab:coverage-summary} distinguishes types, product families, airframe variants and complete parameter records. Here, product families are manufacturer-designation groups; the broader families used for statistical evaluation are defined in Section~\ref{sec:evidence}. Each complete parameter record combines an airframe variant with an applicable engine databank entry and contains all required parameters, including documented approximations.

Table~\ref{tab:coverage} shows one example airframe--engine installation per ICAO type. The variant, engine entry, coefficients, mass and wing area in each row refer to the same installation.

We compared type identifiers with the pycontrails PS parameter and substitution tables dated 28 March 2025, at commit 0d4912374c06~\cite{pycontrailsCoverage2026}. All 68 types in its direct table occur in our inventory. The reconstruction adds 22 type-specific descriptions, of which four already have substitutes: A19N, B74D, C919 and CRJ7. The remaining 18 have neither a direct entry nor a listed substitute. These counts include developmental or discontinued programmes and describe parameter availability rather than traffic coverage or prediction accuracy.


\begin{table}[t]
\caption{Parameter availability and comparison with the pinned pycontrails PS tables. Available variants distinguish airframe versions; complete records link variants to engine entries.}
\label{tab:coverage-summary}
\centering\small
\begin{tabular*}{\columnwidth}{@{\extracolsep{\fill}}lr@{}}
\toprule
Inventory or baseline comparison & Count \\
\midrule
ICAO types & 90 \\
Product families & 82 \\
Available airframe variants & 260 \\
Complete parameter records & 928 \\
Unresolved parameter records & 2 \\
\bottomrule
\end{tabular*}
\par\smallskip
{\footnotesize Status of the 90 reconstructed types in pycontrails\par}
\smallskip
\setlength{\fboxsep}{3pt}
\noindent \colorbox{blue!12}{\parbox[c][10mm][c]{0.28\columnwidth}{\centering\footnotesize\textbf{68}\\ Direct entry}}\hfill \colorbox{orange!12}{\parbox[c][10mm][c]{0.28\columnwidth}{\centering\footnotesize\textbf{4}\\ Substitute only}}\hfill \colorbox{green!12}{\parbox[c][10mm][c]{0.28\columnwidth}{\centering\footnotesize\textbf{18}\\ No entry or substitute}}
\end{table}


\begin{table*}[p]
\caption{One reconstructed installation per ICAO type. Variant and engine identify the selected record. $\mathrm{MTOM}_{\mathrm{ref}}$ and $S_{\mathrm{ref}}$ come from the same record, and design mass is $0.80\,\mathrm{MTOM}_{\mathrm{ref}}$. Coefficients are dimensionless and $\psi_5$ is divided by $10^7$. Types absent from the direct pycontrails table: $\dagger$ has a substitute, $\ddagger$ has none.}
\label{tab:coverage}
\centering
\fontsize{6.6}{7.2}\selectfont
\setlength{\tabcolsep}{1.6pt}
\renewcommand{\arraystretch}{0.64}
\begin{tabular*}{\textwidth}{@{\extracolsep{\fill}}lllrrrrrrrrrrrr@{}}
\toprule
ICAO & Variant & Engine entry & $\psi_0$ & $\psi_2$ & $\psi_3$ & $\psi_4$ & $\psi_5$ & $\psi_6$ & $\psi_7$ & $\tau$ & $\eta_1$ & $\eta_2$ & MTOM (kg) & $S_{\mathrm{ref}}$ (m$^2$) \\
\midrule
A19N$^{\dagger}$ & A319-151N & LEAP-1A24 & 7.70 & 8.24 & 0.530 & 0.712 & 5.96 & 0.752 & 0.888 & 0.179 & 0.357 & 0.401 & 75500 & 122.60 \\
A20N & A320-273N & PW1129G-JM & 7.81 & 8.23 & 0.523 & 0.713 & 5.96 & 0.785 & 0.855 & 0.181 & 0.364 & 0.377 & 79000 & 122.60 \\
A21N & A321-251N & LEAP-1A32 & 8.46 & 8.21 & 0.480 & 0.731 & 6.12 & 0.880 & 0.783 & 0.191 & 0.367 & 0.414 & 93500 & 122.91 \\
A306 & A300B4-605R & CF6-80C2A5-2GE039 & 8.73 & 8.40 & 0.465 & 0.768 & 9.36 & 0.693 & 0.872 & 0.149 & 0.358 & 0.534 & 171700 & 260.00 \\
A30B & A300B4-203 & CF6-50C2-3GE074 & 8.07 & 8.10 & 0.483 & 0.783 & 9.53 & 0.641 & 0.881 & 0.139 & 0.353 & 0.552 & 165000 & 259.99 \\
A310 & A310-204 & CF6-80C2A2-1GE016 & 9.17 & 8.02 & 0.421 & 0.786 & 8.79 & 0.659 & 0.956 & 0.176 & 0.351 & 0.534 & 144000 & 219.00 \\
A313 & A310-322 & JT9D-7R4E1-1PW027 & 9.23 & 8.35 & 0.437 & 0.771 & 8.62 & 0.779 & 0.839 & 0.175 & 0.340 & 0.536 & 164000 & 219.00 \\
A318 & A318-111 & CFM56-5B8/P & 7.72 & 8.16 & 0.515 & 0.745 & 6.23 & 0.586 & 1.092 & 0.169 & 0.329 & 0.516 & 64500 & 122.60 \\
A319 & A319-111-FENCE & CFM56-5B5/P & 8.23 & 8.22 & 0.486 & 0.748 & 6.25 & 0.691 & 0.951 & 0.176 & 0.331 & 0.516 & 76500 & 122.60 \\
A320 & A320-214 & CFM56-5B4 & 7.64 & 8.00 & 0.510 & 0.745 & 6.23 & 0.709 & 0.868 & 0.163 & 0.334 & 0.516 & 78000 & 122.60 \\
A321 & A321-211-FENCES & CFM56-5B3/P & 9.27 & 8.25 & 0.432 & 0.756 & 6.33 & 0.824 & 0.835 & 0.190 & 0.342 & 0.523 & 93500 & 122.91 \\
A332 & A330-243 & Trent772B-60-01P14RR102 & 7.42 & 8.08 & 0.530 & 0.781 & 11.22 & 0.679 & 0.885 & 0.163 & 0.373 & 0.536 & 242000 & 361.63 \\
A333 & A330-343 & Trent772B-60-01P14RR102 & 7.75 & 8.08 & 0.507 & 0.782 & 11.23 & 0.678 & 0.903 & 0.169 & 0.373 & 0.536 & 242000 & 361.63 \\
A338 & A330-841 & Trent7000-72-04P24RR146 & 6.89 & 8.38 & 0.601 & 0.750 & 10.93 & 0.743 & 0.844 & 0.165 & 0.406 & 0.444 & 251000 & 372.36 \\
A339 & A330-941 & Trent7000-72-04P24RR146 & 7.10 & 8.36 & 0.582 & 0.751 & 10.94 & 0.740 & 0.857 & 0.169 & 0.406 & 0.444 & 251000 & 372.36 \\
A342 & A340-211 & CFM56-5C2P-7CM045 & 7.75 & 8.12 & 0.513 & 0.772 & 11.09 & 0.746 & 0.820 & 0.167 & 0.369 & 0.495 & 260000 & 361.63 \\
A343 & A340-313 & CFM56-5C4P-7CM047 & 8.24 & 8.12 & 0.481 & 0.777 & 11.16 & 0.785 & 0.800 & 0.175 & 0.369 & 0.500 & 276500 & 361.63 \\
A345 & A340-541 & RB211-Trent-553-61 & 7.33 & 8.46 & 0.568 & 0.783 & 12.39 & 0.874 & 0.678 & 0.146 & 0.364 & 0.457 & 380000 & 439.40 \\
A346 & A340-642 & Trent-556-61-8RR045 & 7.29 & 8.46 & 0.570 & 0.785 & 12.42 & 0.870 & 0.679 & 0.145 & 0.366 & 0.479 & 380000 & 439.40 \\
A359 & A350-941 & Trent XWB-84 & 6.66 & 8.27 & 0.612 & 0.786 & 12.50 & 0.640 & 0.898 & 0.143 & 0.404 & 0.445 & 283000 & 443.00 \\
A35K & A350-1041 & Trent XWB-97 & 6.59 & 8.29 & 0.617 & 0.797 & 12.98 & 0.675 & 0.829 & 0.135 & 0.404 & 0.467 & 322000 & 464.30 \\
A388 & A380-841 & Trent970-84-01P18RR103 & 6.46 & 8.31 & 0.630 & 0.821 & 18.03 & 0.625 & 0.809 & 0.108 & 0.390 & 0.458 & 575000 & 845.00 \\
AJ27$^{\ddagger}$ & C909-STD & CF34-10A16 & 7.92 & 8.02 & 0.492 & 0.751 & 5.07 & 0.556 & 1.168 & 0.176 & 0.314 & 0.523 & 40500 & 79.86 \\
B37M & B737-MAX-7 & LEAP-1B25 & 7.56 & 8.24 & 0.537 & 0.730 & 6.21 & 0.735 & 0.888 & 0.173 & 0.370 & 0.460 & 80285 & 127.00 \\
B38M & B737-MAX-8 & LEAP-1B25 & 7.91 & 8.24 & 0.512 & 0.730 & 6.21 & 0.756 & 0.880 & 0.179 & 0.370 & 0.460 & 82644 & 127.00 \\
B39M & B737-MAX-9 & LEAP-1B27 & 8.10 & 8.26 & 0.502 & 0.731 & 6.22 & 0.806 & 0.833 & 0.181 & 0.371 & 0.461 & 88314 & 127.00 \\
B3XM & B737-MAX-10 & LEAP-1B28B1 & 8.00 & 8.28 & 0.509 & 0.730 & 6.22 & 0.820 & 0.814 & 0.178 & 0.372 & 0.463 & 89765 & 127.28 \\
B712 & B717-200 & BR700-715C1-30-4BR004 & 8.18 & 7.91 & 0.467 & 0.760 & 5.54 & 0.633 & 0.976 & 0.165 & 0.334 & 0.546 & 54885 & 92.97 \\
B722 & B727-200-ADVANCED & JT8D-15-SMOKE-1PW009 & 7.72 & 7.83 & 0.480 & 0.770 & 7.20 & 0.648 & 0.838 & 0.132 & 0.281 & 0.627 & 95100 & 153.29 \\
B732 & B737-200-ADVANCED & JT8D-15-SMOKE-1PW009 & 8.98 & 7.91 & 0.417 & 0.787 & 5.67 & 0.638 & 1.011 & 0.180 & 0.266 & 0.627 & 58105 & 91.04 \\
B733 & B737-300-BASE & CFM56-3-B1-20K1 & 8.94 & 7.96 & 0.429 & 0.767 & 5.95 & 0.632 & 0.970 & 0.162 & 0.316 & 0.534 & 63276 & 105.40 \\
B734 & B737-400-BASE & CFM56-3-B2-22K1 & 9.13 & 8.00 & 0.421 & 0.768 & 5.95 & 0.678 & 0.913 & 0.164 & 0.319 & 0.534 & 68039 & 105.40 \\
B735 & B737-500-BASE & CFM56-3-B1-20K1 & 8.53 & 7.96 & 0.450 & 0.763 & 5.92 & 0.622 & 0.967 & 0.156 & 0.316 & 0.534 & 61689 & 105.40 \\
B736 & B737-600-BASE & CFM56-7B20 & 7.89 & 7.87 & 0.485 & 0.756 & 6.37 & 0.575 & 1.084 & 0.172 & 0.327 & 0.527 & 65997 & 124.49 \\
B737 & B737-700-BASE & CFM56-7B24 & 8.15 & 7.89 & 0.469 & 0.759 & 6.39 & 0.606 & 1.043 & 0.176 & 0.332 & 0.532 & 70080 & 124.49 \\
B738 & B737-800-BASE & CFM56-7B27 & 8.56 & 7.95 & 0.450 & 0.758 & 6.39 & 0.684 & 0.942 & 0.181 & 0.335 & 0.536 & 79015 & 124.58 \\
B739 & B737-900ER-WINGLETS & CFM56-7B27 & 8.27 & 8.18 & 0.482 & 0.752 & 6.34 & 0.748 & 0.909 & 0.188 & 0.335 & 0.536 & 85139 & 124.60 \\
B742 & B747-200B & JT9D-7Q-1PW025 & 8.19 & 7.92 & 0.462 & 0.812 & 13.86 & 0.695 & 0.759 & 0.132 & 0.334 & 0.539 & 377842 & 511.00 \\
B743 & B747-300 & JT9D-7R4G2-1PW029 & 7.84 & 7.93 & 0.483 & 0.886 & 15.13 & 0.583 & 0.888 & 0.127 & 0.340 & 0.541 & 377842 & 511.00 \\
B744 & B747-400-BASE & CF6-80C2B1F-07P27GE218 & 7.53 & 8.14 & 0.519 & 0.896 & 15.74 & 0.566 & 0.964 & 0.136 & 0.357 & 0.534 & 396893 & 541.16 \\
B748 & B747-8F & GEnx-2B67-P-07P27GE238 & 7.30 & 8.25 & 0.552 & 0.846 & 15.04 & 0.699 & 0.804 & 0.140 & 0.381 & 0.467 & 447696 & 554.00 \\
B74D$^{\dagger}$ & B747-400-DOMESTIC & CF6-80C2B1F-07P27GE218 & 7.39 & 7.77 & 0.504 & 0.882 & 15.06 & 0.431 & 1.170 & 0.125 & 0.354 & 0.534 & 276691 & 511.00 \\
B74F$^{\ddagger}$ & B747-400-FREIGHTER & RB211-524G-1RR010 & 7.39 & 8.13 & 0.528 & 0.900 & 15.81 & 0.561 & 0.965 & 0.134 & 0.349 & 0.553 & 396894 & 541.16 \\
B752 & B757-200-BASE & RB211-535E4-3RR028 & 7.84 & 8.11 & 0.499 & 0.773 & 7.95 & 0.648 & 0.881 & 0.139 & 0.338 & 0.557 & 115893 & 185.25 \\
B753 & B757-300-BASE & RB211-535E4B-5RR039 & 8.30 & 8.16 & 0.473 & 0.762 & 7.83 & 0.713 & 0.821 & 0.145 & 0.338 & 0.557 & 123830 & 185.25 \\
B762 & B767-200ER & CF6-80A-1GE010 & 7.83 & 7.99 & 0.494 & 0.773 & 9.83 & 0.655 & 0.859 & 0.143 & 0.361 & 0.536 & 179169 & 283.35 \\
B763 & B767-300 & CF6-80A2-1GE012 & 8.00 & 7.93 & 0.477 & 0.813 & 10.33 & 0.525 & 1.081 & 0.149 & 0.362 & 0.536 & 158758 & 283.35 \\
B764 & B767-400ER & CF6-80C2B8F-8GE101 & 7.90 & 8.31 & 0.512 & 0.773 & 9.96 & 0.727 & 0.855 & 0.162 & 0.363 & 0.534 & 204116 & 291.00 \\
B772 & B777-200-BASE & Trent877-2RR025 & 7.03 & 8.04 & 0.557 & 0.773 & 12.08 & 0.599 & 0.918 & 0.137 & 0.383 & 0.513 & 247207 & 427.80 \\
B773 & B777-300 & Trent892-2RR027 & 7.36 & 8.11 & 0.536 & 0.783 & 12.23 & 0.707 & 0.791 & 0.141 & 0.379 & 0.520 & 299370 & 427.80 \\
B77L & B777-200LR & GE90-115B-07P27GE240 & 7.36 & 8.45 & 0.562 & 0.812 & 12.68 & 0.764 & 0.798 & 0.156 & 0.372 & 0.492 & 347814 & 427.80 \\
B77W & B777-300ER & GE90-115B-07P27GE240 & 7.44 & 8.43 & 0.555 & 0.812 & 12.68 & 0.772 & 0.792 & 0.158 & 0.372 & 0.492 & 351534 & 427.80 \\
B788 & B787-8 & GEnx-1B64/P2 & 6.32 & 8.01 & 0.624 & 0.821 & 12.04 & 0.555 & 1.014 & 0.146 & 0.403 & 0.443 & 227930 & 377.00 \\
B789 & B787-9-HGW & GEnx-1B74/75/P2 & 6.48 & 8.31 & 0.630 & 0.821 & 12.04 & 0.631 & 0.910 & 0.141 & 0.403 & 0.451 & 259228 & 377.00 \\
B78X & B787-10 & GEnx-1B76/P2 & 6.56 & 8.31 & 0.623 & 0.821 & 12.04 & 0.634 & 0.911 & 0.142 & 0.403 & 0.453 & 260362 & 377.00 \\
BCS1 & A220-100 & PW1524G-04P20PW196 & 8.11 & 7.95 & 0.486 & 0.731 & 5.85 & 0.658 & 1.054 & 0.203 & 0.368 & 0.399 & 63730 & 112.30 \\
BCS3 & A220-300 & PW1521G-3 & 7.24 & 8.16 & 0.559 & 0.718 & 5.75 & 0.757 & 0.878 & 0.180 & 0.364 & 0.391 & 70896 & 112.30 \\
C919$^{\dagger}$ & C919-STD & LEAP-1C28 & 7.79 & 8.21 & 0.521 & 0.721 & 6.19 & 0.692 & 0.947 & 0.175 & 0.366 & 0.410 & 75100 & 129.15 \\
CL30$^{\ddagger}$ & CL300 & AS907-1-1A & 7.29 & 7.75 & 0.511 & 0.779 & 4.10 & 0.370 & 1.599 & 0.153 & 0.285 & 0.554 & 17622 & 48.50 \\
CL35$^{\ddagger}$ & CL350 & AS907-2-1A & 6.99 & 7.73 & 0.535 & 0.771 & 4.06 & 0.395 & 1.568 & 0.168 & 0.288 & 0.554 & 18416 & 48.50 \\
CRJ1$^{\ddagger}$ & CRJ-100-ER & CF34-3A1 & 8.91 & 7.75 & 0.421 & 0.765 & 4.01 & 0.507 & 1.361 & 0.203 & 0.281 & 0.508 & 23133 & 48.33 \\
CRJ2$^{\ddagger}$ & CRJ-200-ER & CF34-3B1 & 8.91 & 7.75 & 0.421 & 0.765 & 4.01 & 0.507 & 1.361 & 0.203 & 0.281 & 0.508 & 23133 & 48.33 \\
CRJ7$^{\dagger}$ & CRJ-700 & CF34-8C5B1 & 8.10 & 7.97 & 0.474 & 0.777 & 4.86 & 0.508 & 1.211 & 0.159 & 0.300 & 0.538 & 34019 & 68.56 \\
CRJ9 & CRJ-900 & CF34-8C5 & 8.31 & 8.00 & 0.465 & 0.768 & 4.89 & 0.566 & 1.145 & 0.175 & 0.303 & 0.539 & 38329 & 71.07 \\
CRJX$^{\ddagger}$ & CRJ-1000 & CF34-8C5A1 & 7.97 & 8.00 & 0.485 & 0.771 & 5.13 & 0.560 & 1.142 & 0.172 & 0.308 & 0.540 & 41640 & 77.39 \\
DC93 & DC-9-32 & JT8D-9-1PW006 & 9.05 & 7.83 & 0.409 & 0.790 & 5.75 & 0.533 & 1.213 & 0.184 & 0.268 & 0.626 & 49895 & 92.90 \\
E135 & ERJ135-ER & AE3007A1-3-TYPE1-6AL011 & 7.79 & 7.67 & 0.475 & 0.750 & 4.05 & 0.422 & 1.405 & 0.156 & 0.273 & 0.541 & 19600 & 51.18 \\
E145 & ERJ145-LR & AE3007A1-TYPE1-6AL005 & 8.34 & 7.73 & 0.446 & 0.754 & 4.07 & 0.481 & 1.267 & 0.163 & 0.275 & 0.541 & 22600 & 51.18 \\
E170 & E170-STD & CF34-8E5 & 8.91 & 8.00 & 0.433 & 0.755 & 4.87 & 0.576 & 1.189 & 0.193 & 0.306 & 0.543 & 38600 & 72.72 \\
E190 & E190-STD & CF34-10E5 & 8.36 & 8.06 & 0.466 & 0.760 & 5.52 & 0.553 & 1.178 & 0.177 & 0.322 & 0.534 & 47790 & 92.53 \\
E195 & E195-STD & CF34-10E5A1 & 8.57 & 8.07 & 0.455 & 0.760 & 5.52 & 0.566 & 1.164 & 0.180 & 0.324 & 0.534 & 48790 & 92.53 \\
E290 & E190-E2 & PW1919G-04P20PW200 & 7.45 & 8.10 & 0.539 & 0.707 & 5.42 & 0.678 & 1.001 & 0.188 & 0.357 & 0.386 & 56400 & 103.00 \\
E295 & E195-E2 & PW1921G-04P20PW201 & 7.73 & 7.92 & 0.509 & 0.723 & 5.54 & 0.719 & 0.984 & 0.209 & 0.361 & 0.392 & 62500 & 103.00 \\
E550$^{\ddagger}$ & Legacy500 & AS907-3-1E-A2 & 7.37 & 7.64 & 0.500 & 0.789 & 3.99 & 0.386 & 1.643 & 0.179 & 0.284 & 0.554 & 17400 & 44.85 \\
E75L & E175-STD-LONG & CF34-8E5 & 9.13 & 7.69 & 0.409 & 0.758 & 4.88 & 0.598 & 1.241 & 0.233 & 0.306 & 0.543 & 40370 & 72.72 \\
E75S & E175-LR-SHORT & CF34-8E5 & 9.08 & 7.97 & 0.423 & 0.763 & 4.92 & 0.590 & 1.167 & 0.197 & 0.306 & 0.543 & 40370 & 72.72 \\
F100$^{\ddagger}$ & F100-620 & TAY Mk620-15 & 7.84 & 8.06 & 0.496 & 0.721 & 5.26 & 0.584 & 1.034 & 0.151 & 0.286 & 0.582 & 45810 & 93.50 \\
F2TH$^{\ddagger}$ & Falcon-2000S & PW308C-BS1289 & 7.53 & 7.75 & 0.497 & 0.769 & 4.07 & 0.415 & 1.555 & 0.181 & 0.281 & 0.557 & 19414 & 49.00 \\
F900$^{\ddagger}$ & Falcon-900EX & TFE731-60 & 8.25 & 7.65 & 0.444 & 0.779 & 4.12 & 0.463 & 1.290 & 0.160 & 0.285 & 0.552 & 22226 & 49.00 \\
FA6X$^{\ddagger}$ & Falcon-6X & PW812D & 7.00 & 7.83 & 0.542 & 0.788 & 5.06 & 0.484 & 1.272 & 0.166 & 0.332 & 0.547 & 35153 & 72.40 \\
FA7X$^{\ddagger}$ & Falcon-7X & PW307A & 7.28 & 7.63 & 0.505 & 0.820 & 5.21 & 0.413 & 1.536 & 0.185 & 0.295 & 0.554 & 31751 & 70.70 \\
FA8X$^{\ddagger}$ & Falcon-8X & PW307D & 7.05 & 7.70 & 0.529 & 0.806 & 5.12 & 0.446 & 1.411 & 0.179 & 0.297 & 0.554 & 33112 & 70.70 \\
G280$^{\ddagger}$ & G280 & AS907-2-1G & 7.47 & 7.75 & 0.499 & 0.773 & 3.92 & 0.414 & 1.479 & 0.162 & 0.283 & 0.554 & 17962 & 45.00 \\
GL7T$^{\ddagger}$ & Global-7500 & Passport20-19BB1A & 6.38 & 7.70 & 0.586 & 0.835 & 6.81 & 0.397 & 1.417 & 0.148 & 0.367 & 0.516 & 52095 & 116.50 \\
GLF5 & GV & BR700-710A1-10-4BR008 & 6.77 & 7.93 & 0.566 & 0.773 & 6.00 & 0.403 & 1.391 & 0.135 & 0.324 & 0.554 & 41050 & 105.58 \\
MD82 & DC-9-82 & JT8D-217C-BASE-1PW018 & 8.44 & 7.91 & 0.449 & 0.770 & 6.16 & 0.631 & 1.028 & 0.183 & 0.293 & 0.611 & 67812 & 112.35 \\
MD83 & DC-9-83 & JT8D-217C-BASE-1PW018 & 8.45 & 7.95 & 0.450 & 0.769 & 6.16 & 0.676 & 0.960 & 0.182 & 0.293 & 0.611 & 72575 & 112.35 \\
MRJ9$^{\ddagger}$ & MRJ90-STD & PW1217G & 7.03 & 8.05 & 0.563 & 0.725 & 5.19 & 0.519 & 1.203 & 0.162 & 0.345 & 0.460 & 39600 & 89.75 \\
RJ1H & AVRO-146-RJ100 & LF507-1F-1TL004 & 9.25 & 8.10 & 0.424 & 0.710 & 4.72 & 0.731 & 0.910 & 0.179 & 0.266 & 0.534 & 46039 & 77.30 \\
SU95$^{\ddagger}$ & RRJ-95B & SaM146-1S17 & 8.05 & 7.79 & 0.468 & 0.766 & 5.29 & 0.578 & 1.086 & 0.175 & 0.303 & 0.549 & 45880 & 83.80 \\
\bottomrule
\end{tabular*}
\end{table*}

\section{Evaluation}
\label{sec:evidence}

The evaluation connects parameter reconstruction to fuel prediction through three comparisons: agreement with published coefficients, efficiency prediction for an excluded aircraft family, and fuel prediction on recorded flights. The first two use published coefficients as references, while the third compares both complete parameter sets against measured fuel flow.

\subsection{Agreement with published parameters}
\label{sec:scoring}

We compared reconstructed parameters with two references (Table~\ref{tab:independent}). The primary comparison used the pinned pycontrails table as a coefficient reference for all 68 direct types~\cite{pycontrailsCoverage2026}, using its drag, design Mach, optimum lift, wave-drag and engine parameters. For the additional $\psi$ coefficients and $\tau$, we used the 53 types in the published aerodynamic table~\cite{poll2024optimum}. These quantities are not directly tabulated for all 68 pycontrails types, so we compared only the available reference values, without reconstructing missing targets.

For each type, we averaged coefficients over its available aircraft--engine configurations, counting identical physical inputs once, then compared the average with the published value. Each type received equal weight in the reported errors; the inventory table instead shows the coefficients of the selected installation. The two unresolved A340-213 records were excluded, while other A342 variants remained in the comparison. Differences in engine, mass and wing configuration therefore contribute to these catalogue comparisons. The differences also include uncertainties in both sets of coefficients. We report fitted quantities, $\eta_1$ and $\psi_1$, separately as descriptive agreement because their errors do not independently test the efficiency regression. The family-excluded evaluation below assesses that fitting procedure.

We measured relative error using mean absolute percentage error (MAPE):
\begin{equation}
\mathrm{MAPE}=\frac{100}{n}\sum_{i=1}^{n}
\left|\frac{\hat y_i-y_i}{y_i}\right|,
\label{eq:mape}
\end{equation}
where $\hat y_i$ is the reconstructed value, $y_i$ the published reference and $n$ the number of comparisons. All references were nonzero. Signed percentage bias used the same average without the absolute value, showing whether predictions were systematically high or low. We also reported the largest absolute percentage error across the evaluated types, to show how far individual cases could depart from the average.

\subsection{Estimating efficiency for an excluded aircraft family}
\label{sec:efficiency-evaluation}

Table~\ref{tab:efficiency} compares three ridge fitting protocols with the mean of the training families.

We evaluated the regression on 66 ICAO types in 21 aircraft families. Families group related airframe designs, so the A320ceo and A320neo series, for example, are held out together.

In each of 21 rounds, we fitted the regression on 20 families and predicted the remaining family. The fixed-formula procedure used the four terms in Section~\ref{sec:efficiency-method}. Global selection chose terms using all 21 families before evaluating the regression. Nested selection chose terms using only the 20 training families~\cite{cawley2010selection}. All three excluded the evaluated family from fitting the regression coefficients; only nested selection also excluded it from choosing the terms.

Both selection procedures considered the same 255 combinations of one to four terms, drawn from the four raw inputs, their logarithms and fuel flow divided by thrust. Selection minimised family-averaged mean absolute error (MAE); nested selection used an additional family-excluded evaluation within the 20 training families. Ridge settings were identical, with ties resolved by fewer terms and then a fixed order. The comparison measures the effect of selecting inputs within this candidate set.

We reported MAE in $\eta_1$ units, MAPE and signed percentage bias. Errors were averaged over input combinations within each type, then over types within each family, then equally across families. A predictor using the training families' mean efficiency provided a baseline. We repeated the analysis with one median input vector per type to test sensitivity to the representation of installations. This also changes the effective regularisation through the number of training observations. The inventory and QAR comparison use the original fitted regression; only this evaluation refits models after excluding each family.

\subsection{Fuel prediction on recorded flights}
\label{sec:qar-method}

The fuel comparison used 100 QAR flights from eight individual A320-214 aircraft, recorded between December 2016 and November 2018. We excluded sharklet variants and required measured mass, airspeed, altitude, vertical speed, temperature, elapsed time and both engine fuel flows. We selected 100 distinct flights with at least 600 valid cruise observations before computing fuel predictions.

For each flight, we evaluated Poll--Schumann with OpenLeaf~\cite{openleaf2026} with two parameter sets for the modelled CFM56-5B4/P: the published coefficients distributed with pycontrails and our reconstruction from ICAO EDB v32 entry 3CM026~\cite{icaoedb32}. The recorded aircraft were equipped with CFM56-5B4/P engines. Both sets used the same recorded states, independent A320-214 geometry ($S_{\mathrm{ref}}=122.6$\,m$^2$), two engines, and EDB idle and take-off fuel-flow values. We retained the published model coefficients with these common inputs. Table~\ref{tab:coverage} shows a base 5B4 installation as the A320 inventory example. A320 references contributed to fitting the reconstructed efficiency model, but QAR fuel measurements were not used for fitting.

Measured total fuel flow was the sum of the two engine channels. Pressure was calculated from recorded pressure altitude using the International Standard Atmosphere; density also used recorded temperature. At each valid observation, OpenLeaf used these conditions, recorded mass and speed to calculate drag and required thrust. Required thrust included the effects of climb or descent and longitudinal acceleration, with a lower bound of zero.

Acceleration was estimated from true airspeed using a quadratic Savitzky--Golay derivative filter with a 241-sample window at 1~Hz. Each continuous valid segment was processed separately; shorter segments used the largest available odd window, with at least three samples. Polynomial interpolation handled segment ends. Recorded airspeed itself was retained in the aerodynamic calculation. Missing observations and time gaps were excluded rather than interpolated. We applied the same limit equations and deterioration correction to both parameter sets: the wave-drag parameter was $J_3=40$, and $\eta_1$ was reduced by 2.5\% to account for deterioration.

We reported errors separately for climb, cruise and descent, using the phase labels supplied with the recordings. Climb included initial climb and descent included approach. Cruise MAPE was the primary metric because this phase best matches the clean aerodynamic configuration used in the calculation. Results outside cruise provide an additional assessment under that approximation, without explicitly modelling changes in flap, landing-gear or speed-brake configuration.

For each flight and phase, we calculated MAPE, signed percentage bias and mean absolute fuel flow error in kg/s. We then averaged the flight-level errors with equal weights. The primary comparison was the change in cruise MAPE, reconstructed minus published, in percentage points. Both runs used identical valid observations, and flights could not be replaced on the basis of prediction errors or failures.

We also integrated measured and predicted fuel flow over the same consecutive valid one-second intervals using the trapezoidal rule. For each flight, fuel error was $100|\hat F-F|/F$, where $\hat F$ and $F$ are predicted and measured fuel mass over those intervals. Reported fuel errors are averages of these per-flight percentages. The combined airborne result excludes takeoff, landing and uncovered intervals; it is not complete trip fuel.

\section{Results}
\label{sec:results}

We first report agreement with published coefficients, then evaluate efficiency prediction for excluded aircraft families. The final comparison assesses the fuel predictions of the complete parameter sets on recorded A320-214 flights.

\subsection{Agreement with published parameters}

Published parameter sets provide benchmarks, but include empirical approximations and calibration~\cite{poll2020part2,poll2024optimum}; the compared aircraft--engine configurations also differ. Agreement therefore does not establish which parameterisation is more accurate. Fuel-prediction performance is assessed against recorded measurements in Section~\ref{sec:qar-results}.

Agreement varied across parameters in both references (Table~\ref{tab:independent}). Across the 68 pycontrails types, MAPE was 7.38\% for the drag parameter $\psi_0$, 3.14\% for design Mach and 5.56\% for the design-optimum lift coefficient. Wave-drag coefficient $J_2$ and the engine Mach exponent $\eta_2$ differed by less than one percent on average. Both $\psi_0$ and the design-optimum lift coefficient had positive bias; their largest absolute percentage errors occurred for B763 and E75L, respectively.

In the 53-type comparison, $\tau$ had the largest MAPE, at 8.98\%, with a maximum error of 34.43\% for CRJ9. Its positive bias indicates that the reconstructed sensitivity of span efficiency to skin friction was generally higher than the published reference. The $\psi_0$ and $\psi_7$ biases were also positive, whereas $\psi_3$ was lower than the published reference on average. Conversely, $\psi_6$ had MAPE 6.15\% but bias close to zero, despite its largest absolute percentage error occurring for B722: errors of opposite signs cancelled in the average. A low bias therefore did not imply close agreement for individual types. These maxima identify individual cases. We did not test whether the largest errors share a common aircraft or engine characteristic.

\begin{table}[t]
\centering
\caption{Agreement of type-average reconstructed parameters with direct references. Errors are percentages, with equal weight per aircraft type. Configuration differences remain. Max. denotes the largest absolute percentage error. Fitted quantities are shown separately as descriptive agreement.}
\label{tab:independent}
\small
\setlength{\tabcolsep}{4pt}
\begin{tabular}{lrrr}
\toprule
Parameter & MAPE & Bias & Max. \\
\midrule
\multicolumn{4}{l}{\textit{pycontrails, 68 types}} \\
$C_{L,\mathrm{DO}}$ & 5.56 & +4.74 & 24.68 \\
$\eta_2$ & 0.62 & +0.09 & 4.75 \\
$J_1$ & 2.62 & -2.39 & 8.47 \\
$J_2$ & 0.23 & -0.19 & 0.97 \\
$M_{\mathrm{EC}}$ & 0.37 & -0.10 & 2.22 \\
$\psi_0$ & 7.38 & +5.59 & 28.44 \\
$\psi_4$ & 3.14 & +0.73 & 12.46 \\
$T_{\mathrm{REC}}$ & 1.20 & -0.52 & 4.89 \\
\midrule
\multicolumn{4}{l}{\textit{Published aerodynamic table, 53 types}} \\
$\eta_2$ & 0.63 & -0.02 & 4.67 \\
$\psi_0$ & 7.87 & +5.93 & 28.43 \\
$\psi_2$ & 1.76 & +1.63 & 5.34 \\
$\psi_3$ & 6.17 & -3.73 & 20.23 \\
$\psi_4$ & 2.75 & +1.11 & 14.93 \\
$\psi_5$ & 2.90 & +0.98 & 10.17 \\
$\psi_6$ & 6.15 & -0.23 & 24.76 \\
$\psi_7$ & 7.83 & +5.76 & 24.84 \\
$\tau$ & 8.98 & +7.33 & 34.43 \\
\midrule
\multicolumn{4}{l}{\textit{Fitted agreement: pycontrails, 68 types}} \\
$\eta_1$ & 2.98 & -0.70 & 11.47 \\
\midrule
\multicolumn{4}{l}{\textit{Fitted agreement: published table, 53 types}} \\
$\eta_1$ & 3.85 & -2.62 & 13.59 \\
$\psi_1$ & 7.67 & -5.70 & 18.65 \\
\bottomrule
\end{tabular}
\end{table}

\subsection{Efficiency estimation and the effect of input selection}
\label{sec:eta1}

The regressions predicted efficiency more accurately than the training families' mean when evaluated on an excluded family (Table~\ref{tab:efficiency}). MAPE was 3.91\% when the evaluated family was excluded from both input selection and fitting, compared with 11.62\% for the mean predictor. The fixed formula, refitted on each training set, gave 3.40\%. Public aircraft and engine characteristics therefore retained predictive value across these families.

Allowing all families to influence input selection gave the same formula and predictions as the fixed-formula procedure. Nested selection retained the fixed predictors in 20 of 21 folds. The A380 fold was the only exception and accounted for the difference in MAPE.

With one median input vector per type, all three procedures selected or used the fixed terms and gave MAPE 3.44\%. The selection effect was therefore concentrated in one family and depended on how installations were represented. These results assess the regression procedures; they do not replace the fitted model used in the inventory.

\begin{table}[t]
\caption{Retrospective efficiency errors on 66 types in 21 aircraft families. Errors are averaged within types, within families, then equally across families. All methods exclude the evaluated family from fitting; nested selection also excludes it from choosing inputs. MAE is in $\eta_1$ units; MAPE and signed relative bias are percentages.}
\label{tab:efficiency}
\centering
\small
\setlength{\tabcolsep}{3pt}
\begin{tabular*}{\columnwidth}{@{\extracolsep{\fill}}lrrr@{}}
\toprule
Method & MAE & MAPE (\%) & Bias (\%) \\
\midrule
\multicolumn{4}{@{}l}{\textit{Distinct input combinations (primary)}} \\
Nested selection & 0.01373 & 3.91 & -0.21 \\
Global selection & 0.01170 & 3.40 & +0.30 \\
Fixed formula & 0.01170 & 3.40 & +0.30 \\
Training-only mean & 0.03839 & 11.62 & +1.75 \\
\midrule
\multicolumn{4}{@{}l}{\textit{Median inputs per type (sensitivity)}} \\
Nested selection & 0.01173 & 3.44 & +0.26 \\
Global selection & 0.01173 & 3.44 & +0.26 \\
Fixed formula & 0.01173 & 3.44 & +0.26 \\
Training-only mean & 0.03839 & 11.62 & +1.75 \\
\bottomrule
\end{tabular*}
\end{table}

\subsection{Fuel prediction on the A320-214 flights}
\label{sec:qar-results}

With the two 5B4/P sets and common inputs, cruise fuel flow MAPE increased from 4.63\% to 5.03\% (+0.41 percentage points), and bias changed from $-2.64$\% to $-3.09$\% (Table~\ref{tab:qar}). Mean cruise MAPE increased for each of the eight aircraft.

Results across phases were mixed. Climb errors increased slightly, while descent fuel flow MAPE decreased despite a larger integrated-fuel error. Over the covered airborne intervals, integrated-fuel error increased from 6.86\% to 8.18\% (+1.32 percentage points). As a sensitivity check, comparing the original published /P set with the base-5B4 reconstruction, each with its original geometry and fuel-flow inputs, gave a small decrease in cruise MAPE (4.68\% to 4.50\%).

\begin{table}[t]
\centering
\caption{Fuel flow and integrated-fuel errors on 100 A320-214 flights, with equal flight weights. P: published 5B4/P coefficients with common inputs. R: reconstructed 5B4/P coefficients. Bias is signed relative fuel flow error. Climb includes initial climb; descent includes approach. Fuel APE is absolute percentage error in integrated fuel over common covered intervals. Noncruise results use the clean-configuration model; takeoff and landing are excluded.}
\label{tab:qar}
\footnotesize
\setlength{\tabcolsep}{3pt}
\begin{tabular}{llrrrr}
\toprule
Phase & Set & MAPE & Bias & MAE & Fuel APE \\
 & & (\%) & (\%) & (kg/s) & (\%) \\
\midrule
Climb & P & 9.60 & -8.24 & 0.138 & 9.31 \\
 & R & 10.55 & -9.44 & 0.152 & 10.61 \\
\addlinespace[2pt]
Cruise & P & 4.63 & -2.64 & 0.031 & 3.06 \\
 & R & 5.03 & -3.09 & 0.034 & 3.49 \\
\addlinespace[2pt]
Descent & P & 27.51 & -5.89 & 0.094 & 16.13 \\
 & R & 27.11 & -12.76 & 0.099 & 21.83 \\
\addlinespace[2pt]
All airborne & P & 11.09 & -4.63 & 0.069 & 6.86 \\
 & R & 11.44 & -6.77 & 0.075 & 8.18 \\
\bottomrule
\end{tabular}
\end{table}

\section{Discussion and Future Work}
\label{sec:discussion}

The reconstruction supplies 22 type-specific descriptions absent from the pinned pycontrails parameter table, including 18 without a listed substitute. This is useful for trajectory studies that would otherwise require substituting a different aircraft type or leaving the type unrepresented. The measured agreement on known types provides a basis for judging the approximations, but accuracy on the added types remains unmeasured. For the four types with substitutes, comparative accuracy would require flight data and a paired test of each substitute and reconstructed set.

The differences in parameter agreement follow distinct calculation paths. The drag scale $\psi_0$ depends on estimated wetted areas and empirical shape corrections, whereas $\tau$ depends on the response of span efficiency to skin friction. Both feed the design-optimum calculation, linking their uncertainties to the lift-to-drag scale $\psi_3$ and optimum-mass scale $\psi_7$. This coupling provides a plausible route for the larger discrepancies in these terms, but does not by itself explain their observed signs or magnitudes. The smaller differences in $J_2$ and $\eta_2$ show closer agreement for those relations, without establishing comparable accuracy for the full aerodynamic model.

Parameter agreement and fuel-prediction accuracy are distinct criteria. Required thrust and propulsion efficiency jointly determine fuel flow (Eq.~\ref{eq:fuel-physical}), so coefficient differences may have compensating effects. Similar aggregate fuel errors establish neither individual coefficient accuracy nor pointwise prediction agreement. Over a restricted operating range, fuel observations may constrain parameter combinations more strongly than individual coefficients: a potential practical identifiability limitation, not demonstrated here. Independent aerodynamic or engine measurements would help assess individual components. The cruise contrast changes sign between the common-input /P comparison and the original configurations, which differ in motorisation and other inputs.

The efficiency analysis supports using public aircraft and engine characteristics to estimate $\eta_1$. It also shows why input selection belongs inside an evaluation intended to assess the full fitting procedure. The nested control retains an advantage over a mean-efficiency predictor, with a selection-related difference concentrated in the A380 fold for the primary representation. These cross-validation results concern the available reference families and the specified candidate inputs. Additional reference families are needed to assess transfer beyond this population.

The QAR comparison provides evidence against measured fuel flow for one aircraft type: the A320-214. These flights do not validate the 90-type inventory or test the added types, and the clean-configuration approximation further limits interpretation outside cruise. Transfer depends on the aerodynamic and turbofan assumptions and the availability of suitable inputs. Other propulsion systems and lighter aircraft require a separate assessment. The QAR assumptions are listed in Section~\ref{sec:qar-method}, including clean aerodynamics, filtered acceleration, and the common 2.5\% efficiency correction. That correction is a modelling assumption. Comparing catalogue errors with ageing-related variability would require longitudinal measurements by configuration and parameter.

Future work will address three needs. First, additional aircraft-family references and more direct geometric measurements will test the reconstruction beyond the present evaluation population. Second, flight comparisons on several types with identified engine configurations will assess fuel accuracy across operating phases. Third, we will use the reconstructed coefficients to parameterise the physical component of hybrid neural ordinary differential equation (Neural ODE) models~\cite{jarry2025neural,jarry2026generation}, and examine how their uncertainty affects trajectory prediction.

\section{Conclusions}

Public airframe and engine information can supply the parameters needed to apply the \PS{} model to a wider range of aircraft descriptions. The method estimates $\eta_1$ when it is unavailable by applying a regression trained on published efficiency scales to public aircraft and engine characteristics. Combined with the physical calculations, this produces an inventory for 90 ICAO type designators. Of the 22 types absent from the direct pycontrails baseline, four already have substitutes. The added descriptions provide explicit inputs for studying those types, whose fuel accuracy remains to be assessed.

The evaluations identify where those inputs remain approximate. Agreement with published physical coefficients varies across quantities, with the largest discrepancies in coupled aerodynamic terms. Public characteristics help estimate efficiency across the evaluated families, although selecting the regression inputs influences the reported accuracy. On the A320-214 cohort, the reconstructed /P coefficients increase cruise fuel flow MAPE by 0.41 percentage points and covered-airborne integrated-fuel error by 1.32 points compared with published /P coefficients under common inputs.

The inventory broadens parameter availability; fuel accuracy across additional types and operating conditions remains to be assessed.

\bibliographystyle{IEEEtran}
\bibliography{references,references_new}

@article{poll2020part1,
  url = {https://doi.org/10.1017/aer.2020.62},
  author  = {Poll, D. I. A. and Schumann, U.},
  title   = {An estimation method for the fuel burn and other performance characteristics of civil transport aircraft in the cruise. Part 1: Fundamental quantities and governing relations for a general atmosphere},
  journal = {The Aeronautical Journal},
  year    = {2020},
  doi     = {10.1017/aer.2020.62}
}

@article{poll2020part2,
  url = {https://doi.org/10.1017/aer.2020.124},
  author  = {Poll, D. I. A. and Schumann, U.},
  title   = {An estimation method for the fuel burn and other performance characteristics of civil transport aircraft during cruise: Part 2, determining the aircraft's characteristic parameters},
  journal = {The Aeronautical Journal},
  year    = {2020},
  doi     = {10.1017/aer.2020.124}
}

@article{poll2024optimum,
  url = {https://doi.org/10.1017/aer.2024.10},
  author  = {Poll, D. I. A. and Schumann, U.},
  title   = {On the conditions for absolute minimum fuel burn for turbofan powered, civil transport aircraft and a simple model for wave drag},
  journal = {The Aeronautical Journal},
  volume  = {128},
  pages   = {1071--1103},
  year    = {2024},
  doi     = {10.1017/aer.2024.10}
}

@article{poll2024engine,
  url = {https://doi.org/10.1017/aer.2024.92},
  author  = {Poll, D. I. A. and Schumann, U.},
  title   = {A simple model for the estimation of turbofan engine performance in all airborne phases of flight},
  journal = {The Aeronautical Journal},
  volume  = {128},
  number  = {1330},
  pages   = {2725--2753},
  year    = {2024},
  doi     = {10.1017/aer.2024.92}
}

@article{poll2025part3,
  url = {https://doi.org/10.1017/aer.2024.141},
  author  = {Poll, D. I. A. and Schumann, U.},
  title   = {An estimation method for the fuel burn and other performance characteristics of civil transport aircraft; Part 3---full flight profile when the trajectory is specified},
  journal = {The Aeronautical Journal},
  year    = {2025},
  doi     = {10.1017/aer.2024.141}
}

@book{obert2009,
  url = {https://research.tudelft.nl/en/publications/aerodynamic-design-of-transport-aircraft/},
  author    = {Obert, Ed},
  title     = {Aerodynamic Design of Transport Aircraft},
  publisher = {IOS Press},
  year      = {2009}
}

@misc{icaoedb32,
  url = {https://www.easa.europa.eu/en/domains/environment/icao-aircraft-engine-emissions-databank},
  author       = {{International Civil Aviation Organization}},
  title        = {ICAO Aircraft Engine Emissions Databank, Version 32},
  year         = {2026},
  howpublished = {Electronic database},
  note         = {Engine identification, rated thrust, bypass ratio, pressure ratio and landing--take-off fuel-flow fields}
}

@article{sun2020openap,
  url = {https://doi.org/10.3390/aerospace7080104},
  author  = {Sun, Junzi and Hoekstra, Jacco M. and Ellerbroek, Joost},
  title   = {OpenAP: An Open-Source Aircraft Performance Model for Air Transportation Studies and Simulations},
  journal = {Aerospace},
  volume  = {7},
  number  = {8},
  pages   = {104},
  year    = {2020},
  doi     = {10.3390/aerospace7080104}
}

@misc{marc_shapiro_2026_19932206,
  author       = {Shapiro, Marc and Engberg, Zeb and Teoh, Roger and Stettler, Marc and Dean, Tom and Abbott, Tristan},
  title        = {{pycontrails}: Python library for modeling aviation climate impacts},
  month        = may,
  year         = {2026},
  publisher    = {Zenodo},
  howpublished = {Zenodo},
  version      = {v0.63.0},
  note         = {Version v0.63.0},
  doi          = {10.5281/zenodo.19932206},
  url          = {https://doi.org/10.5281/zenodo.19932206}
}

@inproceedings{chatterji2011fuel,
  author = {Chatterji, Gano B.},
  title = {Fuel Burn Estimation Using Real Track Data},
  booktitle = {11th AIAA Aviation Technology, Integration, and Operations (ATIO) Conference},
  year = {2011},
  doi = {10.2514/6.2011-6881},
  url = {https://ntrs.nasa.gov/citations/20120002710}
}

@article{schumann2000efficiency,
  author = {Schumann, Ulrich},
  title = {Influence of Propulsion Efficiency on Contrail Formation},
  journal = {Aerospace Science and Technology},
  volume = {4},
  year = {2000},
  doi = {10.1016/S1270-9638(00)01062-2},
  url = {https://elib.dlr.de/9247/}
}

@article{nuic2010bada,
  url = {https://doi.org/10.1002/acs.1176},
  author  = {Nuic, Angela and Poles, Damir and Mouillet, Vincent},
  title   = {{BADA}: An advanced aircraft performance model for present and future {ATM} systems},
  journal = {International Journal of Adaptive Control and Signal Processing},
  volume  = {24},
  number  = {10},
  pages   = {850--866},
  year    = {2010},
  doi     = {10.1002/acs.1176}
}

@article{jarry2024generalization,
  url = {https://doi.org/10.1016/j.trc.2025.105143},
  author  = {Jarry, Gabriel and Dalmau, Ramon and Very, Philippe and Sun, Junzi},
  title   = {On the generalization properties of deep learning for aircraft fuel flow estimation models},
  journal = {Transportation Research Part C: Emerging Technologies},
  volume  = {176},
  pages   = {105143},
  year    = {2025},
  doi     = {10.1016/j.trc.2025.105143},
  note    = {Preprint: arXiv:2410.07717}
}

@article{seymour2020fuel,
  url = {https://doi.org/10.1016/j.trd.2020.102528},
  author  = {Seymour, Kyle and Held, Michael and Georges, Gil and Boulouchos, Konstantinos},
  title   = {Fuel Estimation in Air Transportation: Modeling global fuel consumption for commercial aviation},
  journal = {Transportation Research Part D: Transport and Environment},
  volume  = {88},
  pages   = {102528},
  year    = {2020},
  doi     = {10.1016/j.trd.2020.102528}
}

@book{raymer2018,
  url = {https://doi.org/10.2514/4.104909},
  author    = {Raymer, Daniel P.},
  title     = {Aircraft Design: A Conceptual Approach},
  edition   = {6th},
  publisher = {American Institute of Aeronautics and Astronautics},
  address   = {Reston, VA},
  year      = {2018},
  isbn      = {978-1-62410-490-9},
  doi       = {10.2514/4.104909}
}

@book{torenbeek1982,
  url = {https://doi.org/10.1007/978-94-017-3202-4},
  author    = {Torenbeek, Egbert},
  title     = {Synthesis of Subsonic Airplane Design},
  publisher = {Delft University Press},
  address   = {Delft, The Netherlands},
  year      = {1982},
  isbn      = {978-90-247-2724-7},
  doi       = {10.1007/978-94-017-3202-4}
}

@misc{jarry2025neural,
  author       = {Jarry, Gabriel and Dalmau, Ramon and Olive, Xavier and Very, Philippe},
  title        = {A Neural {ODE} Approach to Aircraft Flight Dynamics Modelling},
  publisher    = {arXiv},
  year         = {2025},
  doi          = {10.48550/arXiv.2509.23307},

  url = {https://doi.org/10.48550/arXiv.2509.23307},
}

@article{jarry2026generation,
  author       = {Jarry, Gabriel and Olive, Xavier},
  title        = {Generation of Vertical Profiles with Neural Ordinary Differential Equations Trained on Open Trajectory Data},
  journal      = {Journal of Open Aviation Science},
  volume       = {4},
  number       = {2},
  year         = {2026},
  month        = mar,
  issn         = {2773-1626},
  publisher    = {TU Delft OPEN Publishing},
  doi          = {10.59490/joas.2026.8457},

  url = {https://doi.org/10.59490/joas.2026.8457},
}

@misc{openleaf2026,
  author = {{EUROCONTROL Aviation Sustainability Unit}},
  title = {{OpenLeaf}: Open-source aviation sustainability suite},
  year = {2026},
  howpublished = {\url{https://github.com/eurocontrol-asu/openleaf}},
  note = {Accessed September 2026}
}

@article{cawley2010selection,
  author={Cawley, Gavin C. and Talbot, Nicola L. C.},
  title={On Over-fitting in Model Selection and Subsequent Selection Bias in Performance Evaluation},
  journal={Journal of Machine Learning Research},
  volume={11}, number={70}, pages={2079--2107}, year={2010},
  url={https://www.jmlr.org/papers/v11/cawley10a.html}
}

@article{sun2020dragpolar,
  url = {https://doi.org/10.1016/j.trc.2020.01.026},
  author = {Sun, Junzi and Hoekstra, Jacco M. and Ellerbroek, Joost},
  title = {Estimating aircraft drag polar using open flight surveillance data and a stochastic total energy model},
  year = {2020},
  doi = {10.1016/j.trc.2020.01.026},
  journal = {Transportation Research Part C: Emerging Technologies},
  volume = {114},
  pages = {391--404}
}

@article{alligier2013mass,
  url = {https://doi.org/10.1016/j.trc.2013.08.006},
  author = {Alligier, R. and Gianazza, D. and Durand, N.},
  title = {Learning the aircraft mass and thrust to improve the ground-based trajectory prediction of climbing flights},
  year = {2013},
  doi = {10.1016/j.trc.2013.08.006},
  journal = {Transportation Research Part C: Emerging Technologies},
  volume = {36},
  pages = {45--60}
}

@inproceedings{chati2017gpr,
  url = {https://doi.org/10.1145/3055004.3055025},
  author = {Chati, Yashovardhan S. and Balakrishnan, Hamsa},
  title = {A {Gaussian} process regression approach to model aircraft engine fuel flow rate},
  year = {2017},
  doi = {10.1145/3055004.3055025},
  booktitle = {Proceedings of the 8th International Conference on Cyber-Physical Systems},
  pages = {131--140}
}

@article{uzun2021physics,
  url = {https://doi.org/10.3390/aerospace8020044},
  author = {Uzun, Mevlut and Demirezen, Mustafa Umut and Inalhan, Gokhan},
  title = {Physics Guided Deep Learning for Data-Driven Aircraft Fuel Consumption Modeling},
  year = {2021},
  doi = {10.3390/aerospace8020044},
  journal = {Aerospace},
  volume = {8},
  number = {2},
  pages = {44}
}

@inproceedings{jarry2024towards,
  author = {Jarry, Gabriel and Delahaye, Daniel and Hurter, Christophe},
  title = {Towards aircraft generic {Quick Access Recorder} fuel flow regression models for {ADS-B} data},
  booktitle = {International Conference on Research in Air Transportation (ICRAT)},
  year = {2024},
  address = {Singapore},
  url = {https://www.icrat.org/previous-conferences/11th-international-conference/papers/}
}

@misc{pycontrailsCoverage2026,
 author={{Contrail Cirrus}},
 title={{pycontrails}: {Poll--Schumann} Aircraft Parameters and Substitutions},
 year={2026},
 url={https://github.com/contrailcirrus/pycontrails/tree/0d4912374c062aeddc443b24a266fe529f4041c3/pycontrails/models/ps_model/static},
 note={Tables dated 28 March 2025; commit 0d4912374c06; accessed 16 September 2026}
}

@techreport{icaoC919Airport,
  author = {{China}},
  title = {The Airport Characteristics of {C919}},
  institution = {International Civil Aviation Organization},
  type = {Working paper},
  number = {AOP/SG/8--WP/26},
  year = {2024},
  note = {Table~1. [Online]. Available: \url{https://www.icao.int/sites/default/files/APAC/Meetings/2024/2024%20AOP-SG-8/3-Working%20Papers/WP26-AI9-THE-AIRPORT-CHARACTERISTICS-OF-C919.pdf}}
}

@techreport{easaA064,
  author = {{European Union Aviation Safety Agency}},
  title = {Type Certificate Data Sheet for {Airbus A318, A319, A320, A321}},
  number = {EASA.A.064},
  institution = {EASA},
  note = {Issue 62, A320 sections 4--5. [Online]. Available: \url{https://www.easa.europa.eu/en/document-library/type-certificates/easaa064}}
}

@techreport{nasaA320Geometry,
  author = {{National Aeronautics and Space Administration}},
  title = {{NASA Technical Publication 20210023843}},
  institution = {NASA},
  note = {Table II, A320 wing geometry. [Online]. Available: \url{https://ntrs.nasa.gov/api/citations/20210023843/downloads/TP-20210023843.pdf}}
}

\end{document}